\documentclass[runningheads]{llncs}
\usepackage{cite}
\usepackage{xcolor,colortbl}
\usepackage{graphicx}
\usepackage{epstopdf}
\graphicspath{{eps/}}
\usepackage{upgreek}
\DeclareGraphicsExtensions{.eps}
\usepackage[cmex10]{amsmath}
\usepackage{amsfonts}
\usepackage{amssymb}
\usepackage{enumerate}
\usepackage{url}
\usepackage{multirow}
\usepackage{siunitx}
\usepackage{scrextend}
\usepackage{tabularx}
\usepackage{fixltx2e}
\usepackage{multirow}
\usepackage{subfigure}
\usepackage{floatrow}

\begin{document}
\title{\textit{S4ND}: Single-Shot Single-Scale Lung Nodule Detection}
\titlerunning{S4ND}
\author{Naji Khosravan \and
Ulas Bagci}
\authorrunning{N. Khosravan et al.}
\institute{Center for Research in Computer Vision (CRCV),\\ School of Computer Science,
University of Central Florida, Orlando, FL.}
\maketitle              % typeset the header of the contribution
\begin{abstract}
The state of the art lung nodule detection studies rely on computationally expensive multi-stage frameworks to detect nodules from CT scans. To address this computational challenge and provide better performance, in this paper we propose S4ND, a new deep learning based method for lung nodule detection. Our approach uses a single feed forward pass of a single network for detection and provides better performance when compared to the current literature. The whole detection pipeline is designed as a single $3D$ Convolutional Neural Network (CNN) with dense connections, trained in an end-to-end manner. S4ND does not require any further post-processing or user guidance to refine detection results. Experimentally, we compared our network with the current state-of-the-art object detection network (SSD) in computer vision as well as the state-of-the-art published method for lung nodule detection (3D DCNN). We used publically available $888$ CT scans from LUNA challenge dataset and showed that the proposed method outperforms the current literature both in terms of efficiency and accuracy by achieving an average FROC-score of $0.897$. We also provide an in-depth analysis of our proposed network to shed light on the unclear paradigms of tiny object detection. 
\end{abstract}

\section{Introduction} 
Successful diagnosis and treatment of lung cancer is highly dependent on early detection of lung nodules. Radiologists are analyzing an ever increasing amount of imaging data (CT scans) every day. Computer Aided Detection (CAD) systems are designed to help radiologists in the screening process. However, automatic detection of lung nodules with CADs remains a challenging task. One reason is the high variation in texture, shape, and position of nodules in CT scans, and their similarity with other nearby structures. Another reason is the discrepancy between the large search space (i.e., entire lung fields) and respectively tiny nature of the nodules. Detection of tiny/small objects has remained a very challenging task in computer vision, which so far has only been solved using computationally expensive multi-stage frameworks. Current sate of art methods for lung nodule detection follow the same multi-stage detection frameworks as in other computer vision areas.

The literature for lung nodule detection and diagnosis is vast. To date, the common strategy for all available CAD systems for lung nodule detection is to use a candidate identification step (also known as region proposal). While some of these studies apply low-level appearance based features as a prior to drive this identification task~\cite{lopez2015large}, others use shape and size information~\cite{krishnamurthy2016automatic}. Related to deep learning based methods, Ypsilantis et al. proposed to use recurrent neural networks in a patch based strategy to improve nodule detection \cite{ypsilantis2016recurrent}. Krishnamurthy et al. proposed to detect candidates using a $2D$ multi-step segmentation process. Then a group of hand-crafted features were extracted, followed by a two-stage classification of candidates \cite{krishnamurthy2016automatic}. In a similar fashion, Huang et al. proposed a geometric model based candidate detection method which followed by a $3D$ CNN to reduce number of FPs \cite{huang2017lung}. Golan et al. used a deep $3D$ CNN with a small input patch of $5\times 20\times 20$ for lung nodule detection. The network was applied to the lung CT volume multiple times using a sliding window and exhaustive search strategy to output a probability map over the volume \cite{golan2016lung}.  

There has, also, been detailed investigations of high-level discriminatory information extraction using deep networks to perform a better FP reduction~\cite{setio2016pulmonary}. Setio et al. used $9$ separate $2D$ convolutional neural networks trained on $9$ different views of candidates, followed by a fusion strategy to perform FP reduction~\cite{setio2016pulmonary}. Another study used a modified version of Faster R-CNN, state of the art object detector at the time, for candidate detection and a patch based $3D$ CNN for FP reduction step \cite{ding2017accurate}. However, all these methods are computationally inefficient (e.g., exhaustive use of sliding windows over feature maps), and often computed in 2D manner, not appreciating the 3D nature of the nodule space. It is worth mentioning that patch based methods are 3D but they suffer from the same computational burdens, as well as missing the entire notion of 3D nodule space due to limited information available in the patches.

\textbf{Our Contributions:} We resolve the aforementioned issues by proposing a completely $3D$ deep network architecture designed to detect lung nodules in a single shot using a single-scale network. To the best of our knowledge, this is the first study to perform lung nodule detection in one step. Specific to the architecture design of the deep network, we make use of convolution blocks with dense connections for this problem, making one step nodule detection computationally feasible. We also investigate and justify the effect of different down-sampling methods in our network due to its important role for tiny object detection. Lastly, we argue that lung nodule detection, as opposed to object detection in natural images, can be done with high accuracy using only a single scale network when network is carefully designed with its hyper-parameters. 

\section{Method}
\begin{figure}[h]
\centering
\includegraphics[scale=0.48]{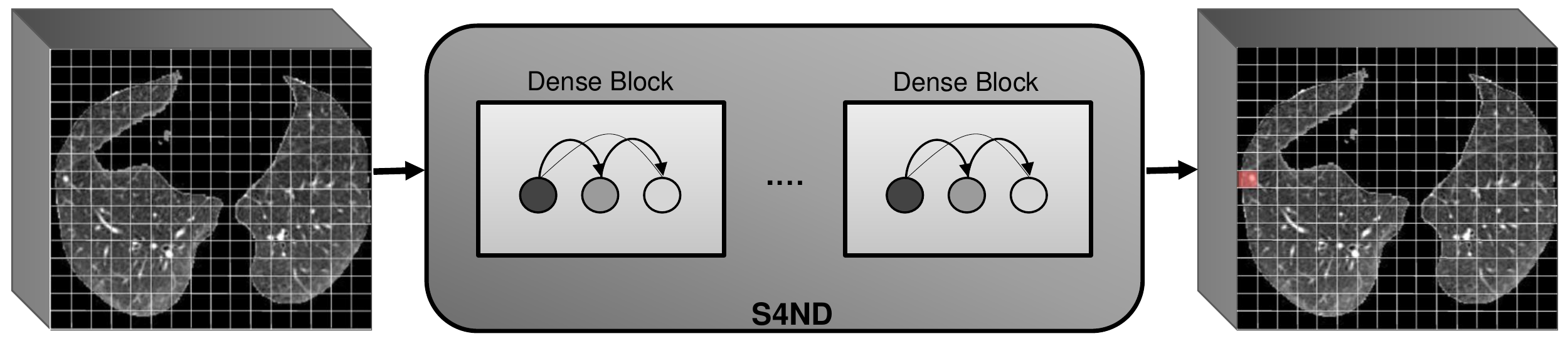}
\vspace{-.4cm}
\caption{Our framework, named S4ND, models nodule detection as a cell-wise classification of the input volume. The input volume is divided by a $16\times 16\times 8$ grid and is passed through a newly designed $3D$ dense CNN. The output is a probability map indicating the presence of a nodule in each cell.\label{fig:sys}}
\end{figure}

Fig.~\ref{fig:sys} shows the overview of the proposed method for lung nodule detection in a single shot. The input to our network is a $3D$ volume of a lung CT scan. The proposed $3D$ densely connected Convolutional Neural Network (CNN) divides the input volume into a grid of size $S\times S\times T$ cells. We model lung nodule detection as a cell-wise classification problem, done simultaneously for all the cells. Unlike commonly used region proposal networks, our proposed network is able to reason the presence of nodule in a cell using global contextual information, based on the whole 3D input volume. 

\vspace{-.6cm}
\vspace{-.8cm}

\subsection{Single-Scale Detection}
As opposed to object detection in natural scenes, we show that lung nodule detection can be performed efficiently and with high accuracy in a single scale. Current literature reports the most frequently observed nodule sizes fall within $3mm$s to $32mm$s~\cite{LUNA16}, most of which are less than $9mm$ and are considered as small (def. American Thoracic Society). Nodules less than $3mm$ in size are the most difficult to detect due to their tiny nature and high similarities to vessels. Based on the statistics of nodule size and the evidence in literature, we hypothesize that a single scale framework with the grid size that we defined ($16\times 16\times 8$ leading to the cell sized of $32\times 32\times 8$ on a volume of size $512\times 512\times 8$) is sufficient to fit all the expected nodule sizes and provide good detection results without the need to increase the algorithmic complexity to multi-scale. This has been partially proven in other multi-scale studies \cite{dou2017multilevel}. 

\vspace{-.4cm}
\subsection{Dense and Deeper Convolution Blocks Improve Detection}
The loss of low-level information throughout a network causes either a high number of false positives or low sensitivity. One efficient way that helps the flow of information in a network and keeps this low-level information, combining it with the high level information, is the use of dense connections inside the convolution blocks. We empirically show that deeper densely-connected blocks provide better detection results. This, however, comes with the cost of more computation. In our experiments we found that dense blocks with $6$ convolution layers provide a good balance of detection accuracy and computational efficiency.

\vspace{-.4cm}
\subsection{Max-Pooling Improves Detection}
As we go deeper in a CNN, it is desired to pick the most descriptive features and pass only those to the next layers. Recently, architectures for object detection in natural images preferred the use of convolutions with stride $2$ instead of pooling~\cite{liu2016ssd}. In the context of tiny object detection, this feature reduction plays an important role. Since our objects of interest are small, if we carelessly pick the features to propagate we can easily lose the objects of interest through the network and end up with a sub-optimal model. In theory, the goal is to have as less pooling as possible. Also, it is desired to have this feature sampling step in a way that information loss is minimized. There are multiple approaches for sampling information through the network. Average pooling, max pooling and convolutions with stride $2$ are some of the options. In our experiments, we showed that max pooling is the best choice of feature sampling for our task as it selects the most discriminative feature in the network. Also, we showed that convolution layers with stride of $2$ are performing better compared to average pooling. The reason is that convolution with stride $2$ is very similar in its nature to weighted averaging with the weights being learned in a data driven manner. 

\vspace{-.4cm}
\subsection{Proposed 3D Deep Network Architecture}
\begin{figure}[h]
\centering
\includegraphics[scale=0.47]{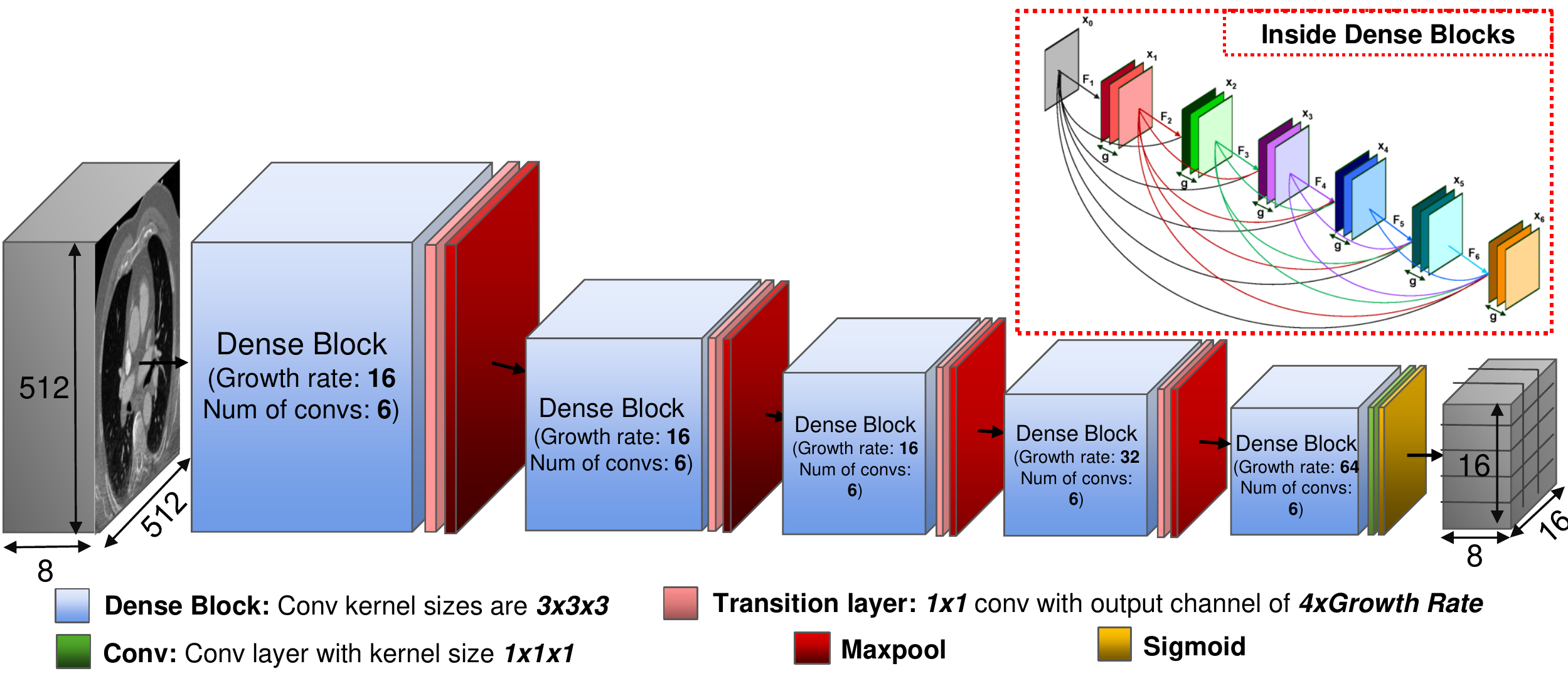}
\caption{Input to the network is a $512\times 512\times 8$ volume and output is a $16\times 16\times 8$ probability map representing likelihood of nodule presence. Our network has $5$ dense blocks each having $6$ conv. layers. The growth rates of  blocks $1$ to $5$ is $16, 16, 16, 32, 64$ respectively. The network has $4$ transition layers and $4$ max-pooling layers. The last block is followed by a convolution layer with kernel size $1\times 1\times 1$ and output channel of $1$ and a sigmoid activation function.\label{fig:architecture}}
\end{figure}

Our network architecture consists of $36$, $3D$ convolution layers, $4$ max-pooling layers and a sigmoid activation function at the end. $30$ of convolution layers form $5$ blocks with dense connections and without pooling, which enhance low-level information along with high-level information, and the remainder form the transition layers. The details of our architecture can be seen in Fig.~\ref{fig:architecture}. The input to our network is $512\times 512\times 8$ and the output is a  $16\times 16\times 8$ probability map. Each cell in the output corresponds to a cell of the original image divided by a $16\times 16\times 8$ grid and decides whether there is a nodule in that cell or not. 

\vspace{-.6cm}

\vspace{-.6cm}
\textbf{Densely connected convolution blocks:} As stated, our network consists of $5$ densely connected blocks, each block containing $6$ convolution layers with an output channel of $g$, which is the growth rate of that block. Inside the blocks, each layer receives all the preceding layers' feature maps as inputs. Fig.~\ref{fig:architecture} (top right) illustrates the layout of a typical dense block. Dense connections help the flow of information inside the network. Assume $x_{0}$ is the input volume to the block and $x_{i}$ is the output feature map of layer $i$ inside the block. Each layer is a non-linear function $F_{i}$, which in our case is a composition of convolution, batch normalization (BN) and rectifier linear unit (ReLU). With dense connections, each layer receives a concatenation of all previous layers' feature maps as input $x_{i}=F_{i}([x_{0},x_{1},...,x_{i-1}])$, where $x_{i}$ is the output feature map from layer $i$ and $[x_{0},x_{1},...,x_{i-1}]$ is the channel-wise concatenation of previous layers' feature maps.

\textbf{Growth rate (GR):} is the number of feature maps that each layer $F_{i}$ produces in the block. This number is fixed for each block but it can change from one block to the other. Assume the number of channels in the input layer of a block is $c_{0}$ and the block has $i$ convolution layers with a growth rate of $g$. Then the output of the block will have $c_{0}+(i-1)g$ channels. 

\textbf{Transition layers:} as can be seen in the above formulations, the number of feature maps inside each dense block increases dramatically. Transition layers are $1\times 1\times 1$ convolution layers with $4\times g$ output channels, where $g$ is the growth rate of previous block. Using a convolution with kernel size of $1\times 1\times 1$ compresses the information channel-wise and reduces the total number of channels throughout the network.

\textbf{Training the network:} The created ground truths for training our network are $3D$ volumes with size $16\times 16\times 8$. Each element in this volume corresponds to a cell in the input image and has label $1$ if a nodule exists in that cell and $0$ otherwise. The design of our network allows for an end-to-end training. We model detection as a cell wise classification of input which is done in one feed forward path of the network in one shot. This formulation detects all the nodules in the given volume simultaneously. The loss function for training our network is weighted cross-entropy defined as:

\vspace{-.6cm}
\begin{equation}
L(Y^{(n)},f(X^{(n)})=\sum_{i=1}^{k_{n}}-y_{i}\log(f(x_{i})),
\end{equation}
where $Y$s are the labels and $X$s are the inputs.

\vspace{-.3cm}
\section{Experiments and Results}
\textbf{Data and evaluation:} To evaluate detection performance of S4ND, we used Lung Nodule Analysis (LUNA16) Challenge dataset (consisting of a total of $888$ chest CT scans, slice thickness$<2.5$ mm, with ground truth nodule locations). For the training, we performed a simple data augmentation by shifting the images in $4$ directions by $32$ pixels. We sampled the 3D volumes for training so that nodules appear in random locations to avoid bias toward location of nodules. We performed $10$-fold cross validation to evaluate our method by following the LUNA challenge guidelines. Free-Response Receiver Operating Characteristic (FROC) analysis has been conducted to calculate sensitivity and specificity~\cite{kundel2008receiver}. Suggested by the challenge organizers, sensitivity at $7$ FP/scan rates (i.e. $0.125, 0.25, 0.5, 1, 2, 4, 8$) was computed. The overall \textit{score} of system (Competition Performance Metric-CPM) was defined as the average sensitivity for these $7$ FP/scan rates. 

\textbf{Building blocks of S4ND and comparisons:} This subsection explains how we build the proposed S4ND network and provides a detailed comparison with several baseline approaches. We compared performance of S4ND with state-of-the-art algorithms, including SSD (single-shot multi-box object detection)~\cite{liu2016ssd}, known to be very effective for object detection in natural scenes. We show that SSD suffers from low performance in lung nodule detection, even though trained from scratch on LUNA dataset. A high degree of scale bias and known difficulties of the lung nodules detection (texture, shape, etc.) in CT data can be considered as potential reasons. To address this poor performance, we propose to replace the convolution layers with \textit{dense} blocks to improve the information flow in the network. Further, we experimentally tested the effects of various down sampling techniques. Table \ref{table:results} shows the results of different network architectures along with the number of parameters based on these combinations. We implemented the SSD based architecture with $3$ different pooling strategies: (1) average pooling (2D Dense Avepool), (2) replacing pooling layers with convolution layers with kernel size $3\times 3$ and stride $2$ (2D Dense Nopool) and (3) max pooling (2D Dense Maxpool). Our experiments show that max pooling is the best choice of feature sampling for tiny object detection as it selects the most discriminating feature in each step. \textit{2D Dense Nopool} outperforms the normal average pooling (\textit{2D Dense Avepool}) as it is in concept a learnable averaging over $3\times 3$ regions of our network, based on the way we defined kernel size and stride. 

\begin{figure}[h]
\centering
\includegraphics[scale=0.28]{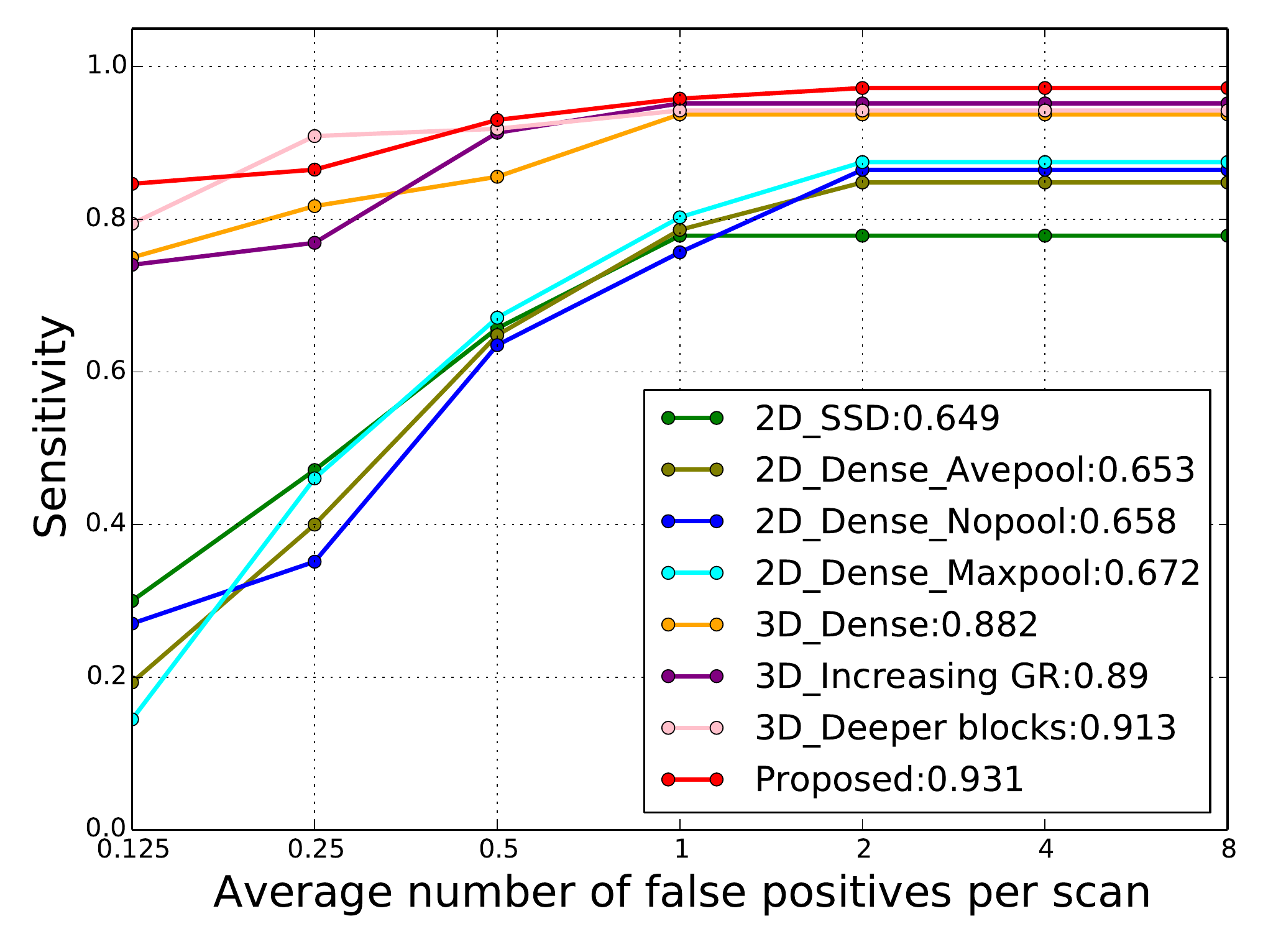}
\includegraphics[scale=0.28]{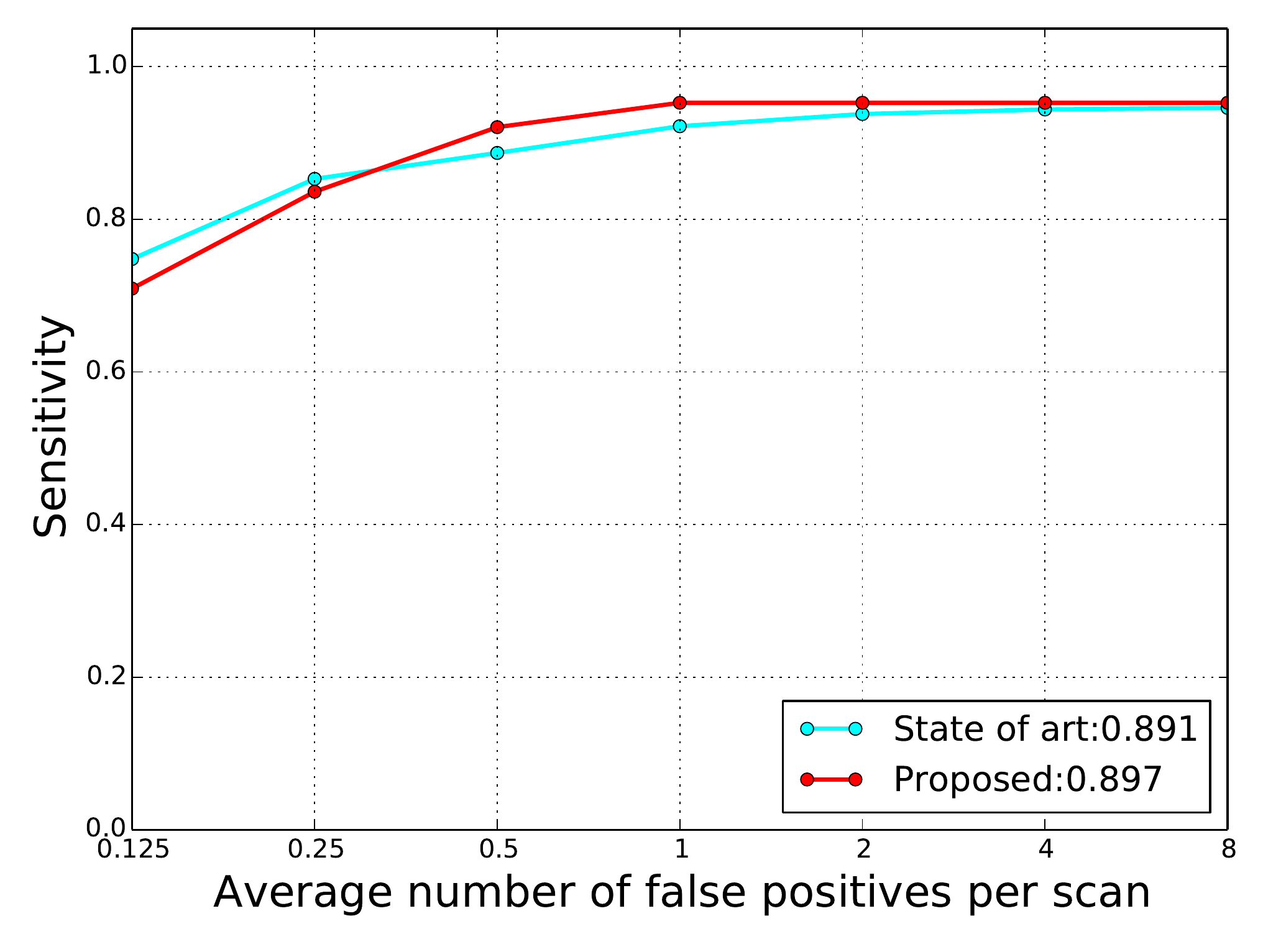}
%\vspace{-.3cm}
\caption{Comparison of base line as well as comparison with the state of the art. Numbers in front of each method in the legend show Competition Performance Metric (CPM). \label{fig:baselinecomparison}}
\end{figure}

\textbf{3D Networks, growth rate (GR), and comparisons:} We implemented S4ND in a completely 3D manner. Growth rate for all the blocks inside the network was initially fixed to $16$ (3D Dense). However, we observed that increasing the growth rate in the last $2$ blocks of our network, where the computational expense is lowest, (from $16$ to $32$ and $64$, respectively) improved the performance of detection (3D Increasing GR in Table \ref{table:results}). Also, having deeper blocks, even with a fixed growth rate of $16$ for all the blocks, help the information flow in the network and improved the results further (3D Deeper Blocks in Table \ref{table:results}). The final proposed method benefits from both deeper blocks and increasing growth rate in its last two blocks. Fig.~\ref{fig:baselinecomparison} (left) shows the FROC comparison of proposed method with the baselines. The 10-fold cross validation results were compared with the current state of the art lung nodule detection method (3D DCNN which is the best published results on LUNA dataset)~\cite{ding2017accurate}. Our proposed method outperformed the best available results both in sensitivity and FROC score, while only using as less as a third of its parameters, and without the need for multi-stage refinements. 

\begin{table}[htb]
\caption{Comparison of different models with varying conditions.}
\vspace{.25cm}

\scalebox{1.1}{
\begin{tabular}{|c|l|c|c|c|}
\hline
\rowcolor{lightgray}
& Model & Sensitivity\% & Num of parameters & CPM\\
\cline{2-5}
\parbox[t]{2mm}{\multirow{8}{*}{\rotatebox[origin=c]{90}{\tiny\textbf{Randomly selected 1-fold}}}}&2D SSD  & 77.8\% & 59,790,787 & 0.649 \\
\cline{2-5}
&2D Dense Avepool & 84.8\% & 67,525,635 & 0.653\\
\cline{2-5}
&2D Dense Nopool & 86.4\% & 70,661,955 & 0.658 \\
\cline{2-5}
&2D Dense Maxpool & 87.5\% & 67,525,635 & 0.672 \\
\cline{2-5}
&3D Dense & 93.7\% & 694,467 & 0.882\\
\cline{2-5}
&3D Increasing GR & 95.1\% & 2,429,827 & 0.890\\
\cline{2-5}
&3D Deeper Blocks & 94.2\% & 1,234,179 & 0.913\\
\cline{2-5}
&Proposed (S4ND) & \textbf{97.2\%} & 4,572,995 & \textbf{0.931}\\
\hline
\parbox[t]{2mm}{\multirow{2}{*}{\rotatebox[origin=c]{90}{\tiny\textbf{10-fold}}}}&3D DCNN \cite{ding2017accurate} & 94.6\% &  11,720,032 & 0.891\\
\cline{2-5}
&Proposed (S4ND) &  \textbf{95.2\%} & 4,572,995 & \textbf{0.897}\\
\hline
\end{tabular}}
\label{table:results}
\end{table}

\vspace{-.3cm}

\vspace{.35cm}
\textbf{Major findings:} (1) We obtained $0.897$ FROC rate in 10-fold cross validation, and consistently outperformed the state of the art methods as well as other alternatives. (2) SSD (the state of the art for object detection in natural images) resulted in the lowest accuracy in all experiments. Proposed S4ND, on the other hand, showed that single scale single shot algorithm performs better and more suited to tiny object detection problem. (3) The proposed method achieved better sensitivity, specificity, and CPM in single fold and 10-fold throughout experiments where S4ND used less than the half parameters of 3D DCNN (current state of the art in lung nodule detection). (4) A careful organization of the architecture helps avoiding computationally heavy processing. We have shown that maxpooling is the best choice of feature selection throughout the network amongst current available methods. (5) Similarly, dense and deeper connections improve the detection rates through better information flow through layers. It should be noted that the runtime of our algorithm for the whole scan, on the test phase, varies from $11~secs$ to $27~secs$ based on the number of slices in the scan on a single NVIDIA TITAN Xp GPU workstation with RAM of $64$ GBs.

\section{Conclusion}
This paper introduces a single-shot single-scale fast lung nodule detection algorithm without the need for additional FP removal and user guidance for refinement of detection process. Our proposed deep network structure is fully 3D and densely connected. We also critically analyzed the role of densely connected layers as well as maxpooling, average pooling and fully convolutional down sampling in detection process. We present a fundamental solution to address the major challenges of current region proposal based lung nodule detection methods: candidate detection and feature resampling stages. We experimentally validate the proposed network's performance both in terms of accuracy (high sensitivity/specificity) and efficiency (less number of parameters and speed) on publicly available LUNA data set, with extensive comparison with the natural object detector networks as well as the state of the art lung nodule detection methods. A promising future direction will be to combine diagnosis stage with the detection.

\bibliographystyle{splncs03}
\bibliography{sssd}

\begin{thebibliography}{10}
\providecommand{\url}[1]{\texttt{#1}}
\providecommand{\urlprefix}{URL }

\bibitem{ding2017accurate}
Ding, J., Li, A., Hu, Z., Wang, L.: Accurate pulmonary nodule detection in
  computed tomography images using deep convolutional neural networks. In:
  International Conference on Medical Image Computing and Computer-Assisted
  Intervention. pp. 559--567. Springer (2017)

\bibitem{dou2017multilevel}
Dou, Q., Chen, H., Yu, L., Qin, J., Heng, P.A.: Multilevel contextual 3-d cnns
  for false positive reduction in pulmonary nodule detection. IEEE Transactions
  on Biomedical Engineering  64(7),  1558--1567 (2017)

\bibitem{golan2016lung}
Golan, R., Jacob, C., Denzinger, J.: Lung nodule detection in ct images using
  deep convolutional neural networks. In: Neural Networks (IJCNN), 2016
  International Joint Conference on. pp. 243--250. IEEE (2016)

\bibitem{huang2017lung}
Huang, X., Shan, J., Vaidya, V.: Lung nodule detection in ct using 3d
  convolutional neural networks. In: Biomedical Imaging (ISBI 2017), 2017 IEEE
  14th International Symposium on. pp. 379--383. IEEE (2017)

\bibitem{krishnamurthy2016automatic}
Krishnamurthy, S., Narasimhan, G., Rengasamy, U.: An automatic computerized
  model for cancerous lung nodule detection from computed tomography images
  with reduced false positives. In: International Conference on Recent Trends
  in Image Processing and Pattern Recognition. pp. 343--355. Springer (2016)

\bibitem{kundel2008receiver}
Kundel, H., Berbaum, K., Dorfman, D., Gur, D., Metz, C., Swensson, R.: Receiver
  operating characteristic analysis in medical imaging. ICRU Report  79(8), ~1
  (2008)

\bibitem{liu2016ssd}
Liu, W., Anguelov, D., Erhan, D., Szegedy, C., Reed, S., Fu, C.Y., Berg, A.C.:
  Ssd: Single shot multibox detector. In: European conference on computer
  vision. pp. 21--37. Springer (2016)

\bibitem{lopez2015large}
Lopez~Torres, E., Fiorina, E., Pennazio, F., Peroni, C., Saletta, M.,
  Camarlinghi, N., Fantacci, M., Cerello, P.: Large scale validation of the m5l
  lung cad on heterogeneous ct datasets. Medical physics  42(4),  1477--1489
  (2015)

\bibitem{LUNA16}
Setio, A.A.A., et~al: Validation, comparison, and combination of algorithms for
  automatic detection of pulmonary nodules in computed tomography images: The
  luna16 challenge. Medical Image Analysis  42(Supplement C),  1 -- 13 (2017),
  \url{http://www.sciencedirect.com/science/article/pii/S1361841517301020}

\bibitem{setio2016pulmonary}
Setio, A.A.A., Ciompi, F., Litjens, G., Gerke, P., Jacobs, C., van Riel, S.J.,
  Wille, M.M.W., Naqibullah, M., S{\'a}nchez, C.I., van Ginneken, B.: Pulmonary
  nodule detection in ct images: false positive reduction using multi-view
  convolutional networks. IEEE transactions on medical imaging  35(5),
  1160--1169 (2016)

\bibitem{ypsilantis2016recurrent}
Ypsilantis, P.P., Montana, G.: Recurrent convolutional networks for pulmonary
  nodule detection in ct imaging. arXiv preprint arXiv:1609.09143  (2016)

\end{thebibliography}
\end{document}